\documentclass[conference]{ieeetran}
\IEEEoverridecommandlockouts
\usepackage{cite}
\usepackage{amsmath,amssymb,amsfonts}
\usepackage{algorithmic}
\usepackage{graphicx}
\usepackage{textcomp}
\usepackage{xcolor}
\def\BibTeX{{\rm B\kern-.05em{\sc i\kern-.025em b}\kern-.08em
    T\kern-.1667em\lower.7ex\hbox{E}\kern-.125emX}}

\newcommand{\comment}[1]{}

\begin{document}
\title{DEEPGONET: Multi-label Prediction of GO Annotation for Protein from Sequence Using Cascaded Convolutional and Recurrent Network}

\author{
\IEEEauthorblockN{Sheikh Muhammad Saiful Islam}
\IEEEauthorblockA{\textit{Department of Pharmacy} \\
	\textit{Manarat International University}\\
	Dhaka, Bangladesh \\
	saifulislam@manarat.ac.bd}

\and
\IEEEauthorblockN{Md Mahedi Hasan}
\IEEEauthorblockA{\textit{Institute of Information and Communication Technology} \\ 
	\textit{Bangladesh University of Engineering and Technology}\\
	Dhaka, Bangladesh \\
	mahedi0803@gmail.com}

}

\comment{
} 

\maketitle

\begin{abstract}
The present gap between the amount of available protein sequence due to the development of next generation sequencing technology (NGS) and slow and expensive experimental extraction of useful information like annotation of protein sequence in different functional aspects, is ever widening, which can be reduced by employing automatic function prediction (AFP) approaches. Gene Ontology (GO), comprising of more than $40,000$ classes, defines three aspects of protein function names Biological Process (BP), Cellular Component (CC), Molecular Function (MF). Multiple functions of a single protein, has made automatic function prediction a large-scale, multi-class, multi-label task. In this paper, we present DEEPGONET, a novel cascaded convolutional and recurrent neural network, to predict the top-level hierarchy of GO ontology. The network takes the primary sequence of protein as input which makes it more useful than other prevailing state-of-the-art deep learning based methods with multi-modal input, making them less applicable for proteins where only primary sequence is available. All predictions of different protein functions of our network are performed by the same architecture, a proof of better generalization as demonstrated by promising performance on a variety of organisms while trained on \textit{Homo sapiens} only, which is made possible by efficient exploration of vast output space by leveraging hierarchical relationship among GO classes. The promising performance of our model makes it a potential avenue for directing experimental protein functions exploration efficiently by vastly eliminating possible routes which is done by the exploring only the suggested routes from our model. Our proposed model is also very simple and efficient in terms of computational time and space compared to other architectures in literature. Our source code is available in https://github.com/saifulislampharma/go\_annotator. 
\end{abstract}


\section{Introduction}
With ever increasing application of next generation sequencing techniques and concomitant decrease in sequencing, millions of protein sequence can be generated within short time and low cost~\cite{Cao1}. Though information about the protein sequences facilitates advancement in many applications like phylogenetics and evolutionary biology, knowledge of the proteins’ functions is required to elucidate the nature and behavior of living systems as well as for more important application such as biomedical and pharmaceutical ones~\cite{Radiv, Cao2}. Assigning functions to protein is challenging and in vitro or in vivo experiments is generally employed~\cite{Costan}. There is an obvious gap between rapid increasing amount of novel protein sequences and experimental functional annotation of proteins, is ever widening.

Computational prediction of protein functions is very promising to address the challenge of annotating function to protein~\cite{Radiv}. A predictive pipeline, capable of predicting function from primary sequence of protein, can form an manageable subset of high confidence candidates by filtering astronomically large data set, which would ultimately be validated by experimental tools, is greatly sought.

Gene ontology (GO) is a major project undertaken to standardize the representation of gene and gene product attributes across all species~\cite{consort1}. GO annotation is divided into three domains, Biological process (BP), Molecular function (MF), cellular component (CC). GO uses controlled vocabulary which is called term to represent above mentioned aspects. The terms are hierarchically represented, where ancestor represent an general concept while the descendant indicates more specific concept. Currently there are more than $40,000$ GO terms presented in GO ontology. 

There are some significant challenges for computational prediction of protein function. One of the challenges is that protein sequence, structure and function are related to each other in a complex way~\cite{alberts}. Before using protein structure to predict protein function, we need protein structure with sufficient quality which require a considerably large endeavor. Huge and intricate output space for classifying protein function constitute second largest challenge. 

Gene Ontology (GO)~\cite{consort1} has more than $40,000$ terms also known as classes with complex hierarchical relationship among them. For this hierarchical representation, protein has to be assigned to multiple classes. Additionally, numerous proteins have several function, which makes the protein function classification a multi-label, multi-class problem. But, present prevailing state-of-the-art deep learning based GO annotation prediction model do not consider hierarchical relationship among the GO term explicitly. As a result it makes difficult for the model to explore the vast output space efficiently as there is not sufficient samples for this task in present GO ontology dataset.  

In this paper, we propose a novel deep learning method to predict top-level of hierarchy of GO annotation function of protein from protein primary sequence only. We extract local features of amino acid by using several convolutional layers with different kernel size which is followed by bidirectional long short term recurrent neural network (Bi-GRU) to extract global pattern in amino acid. These two different features, local and global, are then combined and finally fed into densely connected layers on top of followed by a classifier which predicts the top-level hierarchical classes for each domain in GO ontology.

We demonstrated that our model has state-of-the-art performance in predicting GO classes for proteins from different species, and performs particularly well in domain of cellular location prediction of proteins. The prime contributions of our paper are as follows: \\
1. Our model can efficiently explore the enormous output space by considering hierarchical relationship among GO classes. \\
2. Our network uses the same model for predicting classes on three domains in GO ontology, which proves the better generalization of our model. \\
3. Our model takes no external information for any protein which is a significant because there is not same external information available for every protein, making our model perfectly suitable for prediction, if there is only sequence available for a protein, which is most frequent case for proteins. 

The rest of the paper is organized as follows: section~\ref{related_work} discusses related work, while section~\ref{PM} describes our proposed cascaded convolutional and recurrent network along with the steps regrading data preprocessing and network architecture for protein function predictions. Experiments and evaluations are presented in section~\ref{er}. Finally, we summarize our results in section~\ref{con}.

\section{Related Work} \label{related_work}
Protein function prediction is inherently a multi-label classification problem as single protein can perform different functions in the biological system. Over the last few decades, various computational methods had been tried for this problem~\cite{Watson,Friedberg,Redfern}, which can be broadly categorized into the following methods.

Basic Local Alignment Search Tool (BLAST)~\cite{Altschul} is the most used method for protein function prediction. Databases of experimentally determined functions of protein are queried against target protein, and function is assigned to target protein on the basis of most homologous proteins in the databases. The methods of Gotcha~\cite{Martin2}, OntoBlast~\cite{Zehetner}, and Goblet~\cite{Groth} are also in this category.

The second category consists of network based methods. Majority of the methods in this category, leverages protein\--protein interaction (PPI) networks for predicting protein’s function with underlying assumption of interacted proteins having similar functions~\cite{Hishigaki,Chua,Zeng4}. Beside this, other kind of networks such as gene\--gene interaction network, and domain co-occurrence networks are used in predicting protein’s function~\cite{Wang3, Wan2}.

The machine learning based methods which infer function to a protein sequence without input from any database or other sources, constitute the third category. Most of the existing machine learning methods perform model training by using generated features from the sequences of protein, after which it is used to predict protein function from an input sequence~\cite{Zhang1,Pan1}. Protein sequence, protein secondary structure, hydrophobicity, sub-cellular location, solvent accessibility, etc., are most frequent used features in these methods. But with exception of protein sequence, vast majority of features are predicted from the protein sequence, so protein function prediction can be more erroneous.

Employing deep learning in multiple layers representation and abstraction of data, has revolutionized many fields including many branches of Bioinformatics, though, not very common in tackling protein function prediction, some deep learning based method are tested~\cite{Cao3,Frasca,Kulmanov}. In~\cite{Kulmanov}, multi-modal input consists of primary sequence and protein-protein interaction network was used, while in~\cite{Cao3} protein’s primary sequence was mapped to GO term using neural machine translational model. Authors in~\cite{Frasca} employed task dissimilarity along with task similarity for better prediction of rare classes.

\section{Proposed Method} \label{PM}
\begin{figure}
	\centerline{\includegraphics[width = 85mm]{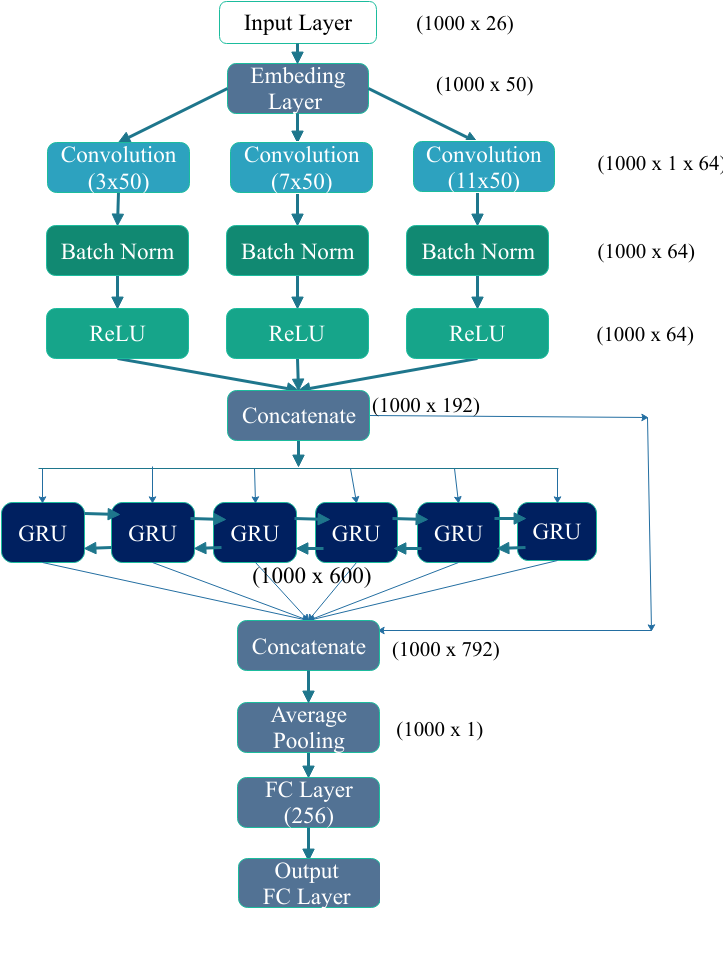}}
	\caption{Proposed architecture of our DEEPGONET network. The Input of our network is the matrix of one-hot encoded amino acid of primary sequence of a protein. Thereafter, an embedding layer is used to produce a dense representation of input feature space from sparse one-hot encoded features. Three sets of local features are extracted in parallel by three convolutional layers with kernel size of 3x50,7x50,11x50. The concatenated local features are fed into a BiGRU layer with 300 neurons to extract global features. The local and global features are concatenated and spatially averaged to reduce dimension of the feature space. Thereafter, a fully connected layer with $256$ neurons followed by another final fully connected layer with sigmoid activation are used to predict top-level of GO Ontology hierarchy. The number of neurons in the final output dense layer varies with 33,22,16 in the prediction of BP, CC, MF domains respectively in three different models. }
	\label{fig_1}
\end{figure}

\subsection{Data Preprocessing}
Firstly, we converted primary protein sequence of amino which is represented by one character assigned by UNIPAC, into 26 dimensional one-hot vector. These $26$ type of amino acid includes $20$ different types of amino acids along with the existing derivatives of these $20$ amino acid. For each GO domain, we derived the ancestors in the top-level of hierarchy and mapped every GO term to one or more ancestors in the top-level of respective domain. This top-level ancestors constitute our target set. The top-level ancestor has $33, 22, 16$ term, for BP, CC, MF respectively. For a protein, we converted each of it's top-level ancestor to form our final target vector using dictionary, consisting of top-level terms in respective GO domain.

\subsection{Network Architecture}

As illustrated in Fig.~\ref{fig_1}, our proposed DEEPGONET has four parts, one feature embedding layer with 50 neurons, followed by three parallel multi-scale convolutional layers, followed by one bidirectional gated recurrent unit (BiGRU)~\cite{Schuster, Cho} layer with $300$ neurons and on top it, one fully connected dense layer and one output dense layer with sigmoid activation. The input matrix of our network is comprised of one-hot encoded amino acid sequence of protein. Since, the sparse one-hot encoding does not provide sufficient generalization, feature embedding layer is used to transform sparse one-hot encoded feature vectors into denser feature vectors. The resulting dense features are then fed into CNN layers with kernel of three different sizes 3 x 50, 7 x 50, 11 x 50 to extract local features in parallel. Thereafter, these local features are concatenated and fed into BiGRU layers to extract global features. Spatial average of the concatenated local and global feature maps extracted by CNN and BiGRU respectively are done by adding a average pooling layer. Thereafter, a fully connected layer of $256$ neurons followed by a sigmoid activated final fully connected layer performs top-level of GO hierarchy classification.

\subsection{Multi-Label Joint Feature Learning}
Since, a protein has more than one member in top-level of GO hierarchy, we went for multi-label binary classification for this work. For predicting the top-level of GO hierarchy, our networks produced  $33, 22, 16$ dimensional output vector for BP, CC, MF GO ontology domains respectively. Each output vector index $i$ was set to $1$ when the respective member of the top-level hierarchy is a parent of current protein, otherwise set to 0. So, our prediction task became multi-label binary classification problem for which we used sigmoid activation and multi-label binary classification loss shown in \eqref{multi_task}.

\begin{equation}
-\frac{1}{m}\sum_{i = 1}^{m}\sum_{j = 1}^{k}({y_j^{(i)}}{\log{\widehat{y}_j^{(i)}}} + {(1 - y_j^{(i)})} {\log{(1 - \widehat{y}_j^{(i)}})})   \label{multi_task}
\end{equation}

We evaluated the performance of our models with two measures. The first one is $F_{1}$-score using precision and recall for proteins. 

\begin{equation}
	F_{1} = 2 * \frac{Precision * Recall}{Precision + Recall} \label{f1}
\end{equation}

\begin{equation}
	MCC = \frac{(TP * TN) -  (FP * FN)} {\sqrt{(TP+FP) (TP+FN) (TN+FP) (TN+FN)}} \label{mcc}
\end{equation}
The second one is Mathew correlation coefficient (MCC)~\cite{Boughorbel}, which is the state-of-the art performance evaluation metric in case of classes which have very different sample size. Equation \eqref{f1} and \eqref{mcc} calculates $F_{1}$ and $MCC$ respectively. Since, our experiment of GO ontology dataset is a multi-label binary classification problem where each member of top-level have very different number of true positive samples,  $F_{1}$-score and $MCC$ are two perfect metrics to evaluate.  

\subsection{Training}
\begin{table}
	\caption{Hyperparameters Setting of Proposed Network}
	\begin{center}
		\begin{tabular}{ccc}
			\hline
			\textbf{Dataset}& \textbf{Hyperparameter}& \textbf{Value} \\ [0.4mm]
			
			\hline \rule{0pt}{3ex}
			Biological Process (BP) & \multicolumn{2}{c}{
				\begin{tabular}{cc}
					Optimizer & Adam\\ \hline
					Learning rate & 1e-5\\ \hline
					Mini-batch size  & 100\\ \hline
					Epochs & 48\\  [0.4mm]
				\end{tabular}		
			}\\ 
		
			\hline \rule{0pt}{3ex}
			Cellular Component (CC)& \multicolumn{2}{c}{
				\begin{tabular}{cc}
					Optimizer & Adam\\ \hline
					Learning rate & 1e-5\\ \hline
					Mini-batch size  & 100\\ \hline
					Epochs & 128\\  [0.4mm]
				\end{tabular}		
			}\\ 
			
			\hline \rule{0pt}{3ex}
			Molecular Function (MF) & \multicolumn{2}{c}{
				\begin{tabular}{cc}
					Optimizer & Adam\\ \hline
					Learning rate & 1e-5\\ \hline
					Mini-batch size  & 100\\ \hline
					Epochs & 155\\  [0.4mm]
				\end{tabular}		
			}\\ \hline
		
		\end{tabular}
		\label{model_training}
	\end{center}
\end{table}

To train each of the three models of our DEEPGONET network we did the following: we used Adam~\cite{Kingma} optimizer function, which is known to work very well for training on recurrent neural networks (RNNs). We used multi-label binary classification loss \eqref{multi_task} as our objective function. We tried several learning rates and found out that the best initial learning rate is $10^{-5}$ for our experiment. As the training progress, we reduce our learning rates further. Our network shows some overfitting. We used dropout with rate of $0.5$ on the output of first fully connected layer to combat overfitting. We also implemented batch normalization layer~\cite{Ioffe} before the recurrent layer to reduce internal co-variance shift. ReLU was used as activation function for fully connected layer. The batch size was set to 100. The entire deep network is trained on a single NVIDIA Tesla K40 GPU with 11GB memory. It takes about three hours to train our model. In the testing stage, one sample takes on average about 7ms.We took 90\% of data for the training and rest for the validation purpose. The summary of our training hyperparameters is shown in Table ~\ref{model_training}.

\subsection{Post-processing}
\begin{figure}
	\centerline{\includegraphics[width= 85mm]{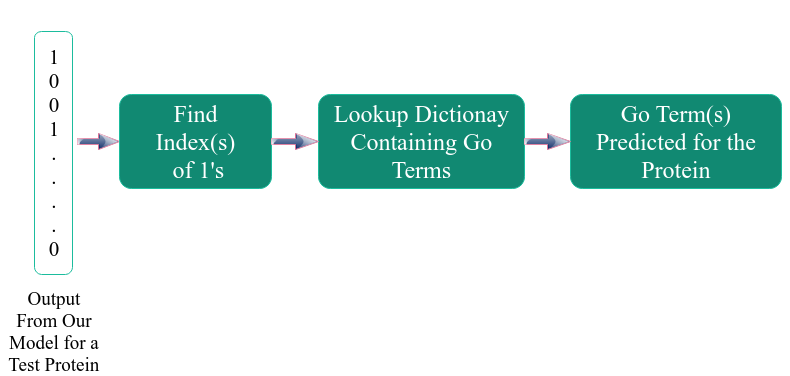}}
	\caption{Output prediction scheme of our proposed DEEPGONET.}
	\label{fig_2}
\end{figure}

After training we need to output the predicted members of top-level hierarchy of GO ontology in three different domains. For prediction, we used key-value dictionary. As depicted in the Fig.~\ref{fig_2}, from the output of any of three models, we extracted the index of 1's in the output vector. We used the index value(s) of 1's to get the GO term label of top-level members in hierarchy.

\section{Experimental Results} \label{er}
\subsection{Dataset}
We used largest publicly available GO annotation database form GO ontology consortium for protein annotations. GO annotation uses controlled vocabulary for defining protein’s annotations. It uses three domains for ontology purpose. It has three set of predefined ontology terms for representing biological process, cellular component, and molecular function. 

For annotation, we used reviewed protein sequences with GO annotations for \textit{Homo sapiens} from SwissProt’s~\cite{Boutel}, a curated protein sequence database, downloaded on $08^{th}$ July 2017 from https://www.ebi.ac.uk/QuickGO/. This dataset contains $647,006$ annotations. We included proteins with experimentally determined annotations, excluding automatic annotation, annotation without biological data, and annotation with unsupported biological data, resulting in total $403,781$ annotation. 

For annotation terms, Gene Ontology (GO)~\cite{consort1}, downloaded on $08^{th}$ July 2017 from http://geneontology.org/page/download-ontology in $OBO$ format was used. In this version of ontology term dataset there are $44,683$ term out of which $1,968$ are obsolete. GO consists of three domains which are CC, BP, MF, each having $3907$, $10161$, $28647$ terms respectively.

\begin{table}
	\caption{Number of Protein Annotation in Different Domains}
	\begin{center}
		\begin{tabular}{p{1.5cm}p{2cm}p{4cm}}
			\hline
			\textbf{GO Domain}& \textbf{Experimentally Determined Annotation}& \textbf{Not Experimentally Determined Annotations} \\ [0.4mm]
			
			\hline 
			\rule{0pt}{3ex}
			$\quad$BP &  $\quad$$103,094$ &  $\quad\qquad\qquad$$99,292$ \\ [0.4mm]
			
			\hline
			\rule{0pt}{3ex}
			$\quad$CC & $\quad$$133,549$ &  $\quad\qquad\qquad$$96,941$ \\ [0.4mm]

			\hline
			\rule{0pt}{3ex}
			$\quad$MF & $\quad$$144,853$ &  $\quad\qquad\qquad$$59,6682$ \\ [0.4mm]
			\hline
		\end{tabular}
		\label{gene_annotation}
	\end{center}
\end{table}

For classification purpose, we splitted these total $403,781$ annotations according to BP, CC, MF domains. Resulting three dataset are described in the Table~\ref{gene_annotation}. We downloaded all the protein sequences from https://www.uniprot.org/.

\begin{table}
	\caption{Aggregated BP, CC, MF Dataset}
	\begin{center}
		\begin{tabular}{p{1.5cm}p{3cm}p{3cm}}
			\hline
			\textbf{GO Domain}& \textbf{Number of Sample before Aggregation of GO Term for Each Protein}& \textbf{Number of sample After Aggregation of GO Term for Each Protein} \\ [0.4mm]
			
			\hline 
			\rule{0pt}{3ex}
			$\quad$BP &  $\qquad\qquad$$103,094$ &  $\qquad\qquad$$14,536$ \\ [0.4mm]
			
			\hline
			\rule{0pt}{3ex}
			$\quad$CC & $\qquad\qquad$$133,549$ &  $\qquad\qquad$$15,740$ \\ [0.4mm]
			
			\hline
			\rule{0pt}{3ex}
			$\quad$MF & $\qquad\qquad$$144,853$ &  $\qquad\qquad$$15,446$ \\ [0.4mm]
			\hline
		\end{tabular}
		\label{go_ontology}
	\end{center}
\end{table}

Since, three dataset are organized in such a way that in each row one protein ID is assigned to one GO term. For each dataset, we aggregated all GO terms for a single protein along with it's primary sequence. The resulted three dataset are summarized in the Table~\ref{go_ontology}

\subsection{Experimental Result on Aggregated Biological Process, Cellular Component a Molecular Function Dataset}
\begin{table}
	\caption{Performance Evaluation of Our Proposed Models In Three Different Domains Of GO Ontology.}
	\begin{center}
		\begin{tabular}{ccc}
			\hline
			\textbf{Network}& \textbf{$F_1$}-score& \textbf{$MCC$} \\ [0.4mm]
			
			\hline 
			\rule{0pt}{3ex}
			Biological Process(BP) &  $\quad$$0.590$ &  $\quad$$0.511$ \\ [0.4mm]
			
			\hline
			\rule{0pt}{3ex}
			Cellular Component (CC) & $\quad$$0.763$ &  $\quad$$0.700$ \\ [0.4mm]
			
			\hline
			\rule{0pt}{3ex}
			Molecular Function (MF) & $\quad$$0.731$ &  $\quad$$0.706$ \\ [0.4mm]
			\hline
		\end{tabular}
		\label{result}
	\end{center}
\end{table}

\begin{figure}
	\centerline{\includegraphics[width= 75mm]{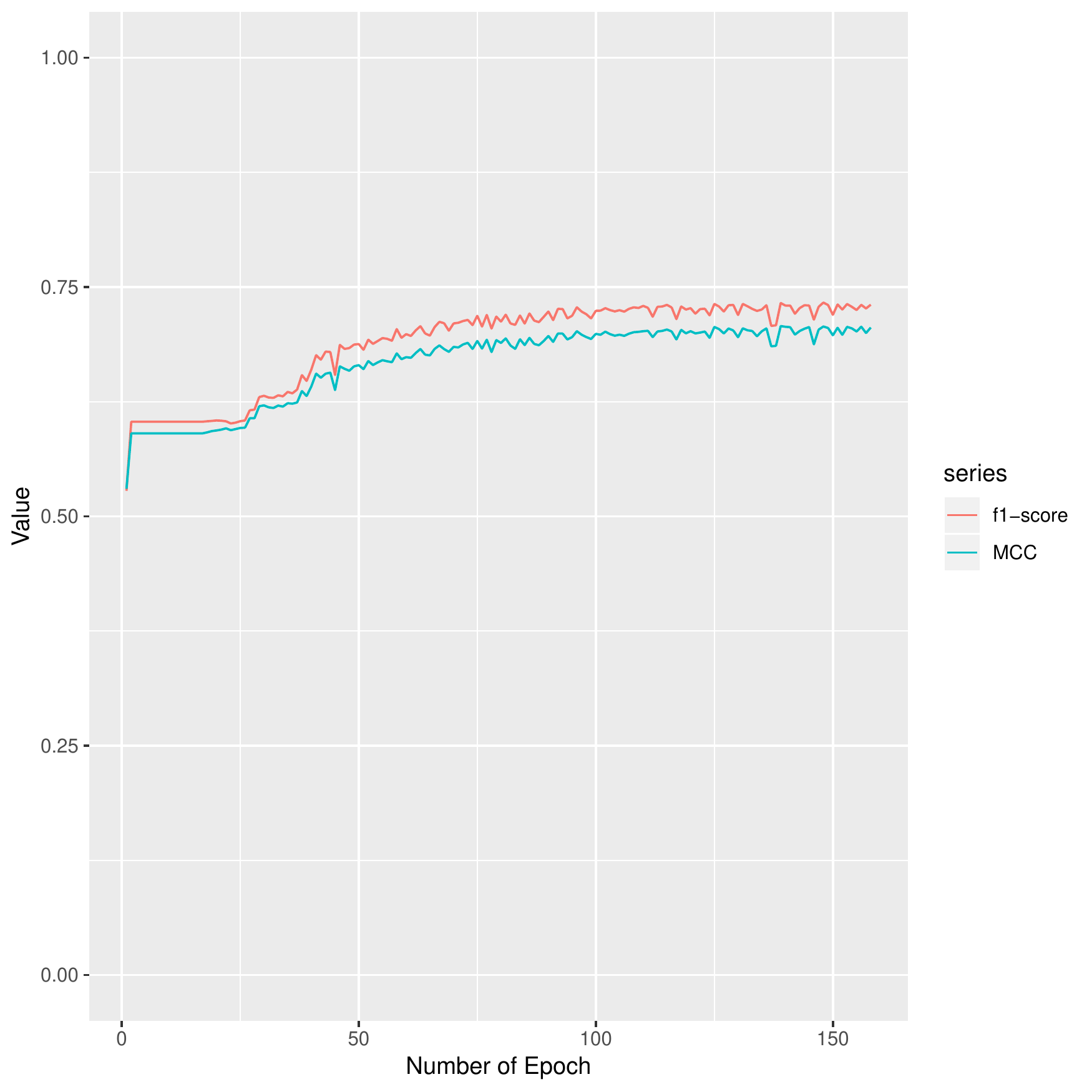}}
	\caption{Performance evaluation of our proposed method on different metric in Molecular Function (MF) dataset.}
	\label{fig_3}
\end{figure}
\begin{figure}
	\centerline{\includegraphics[width = 75mm]{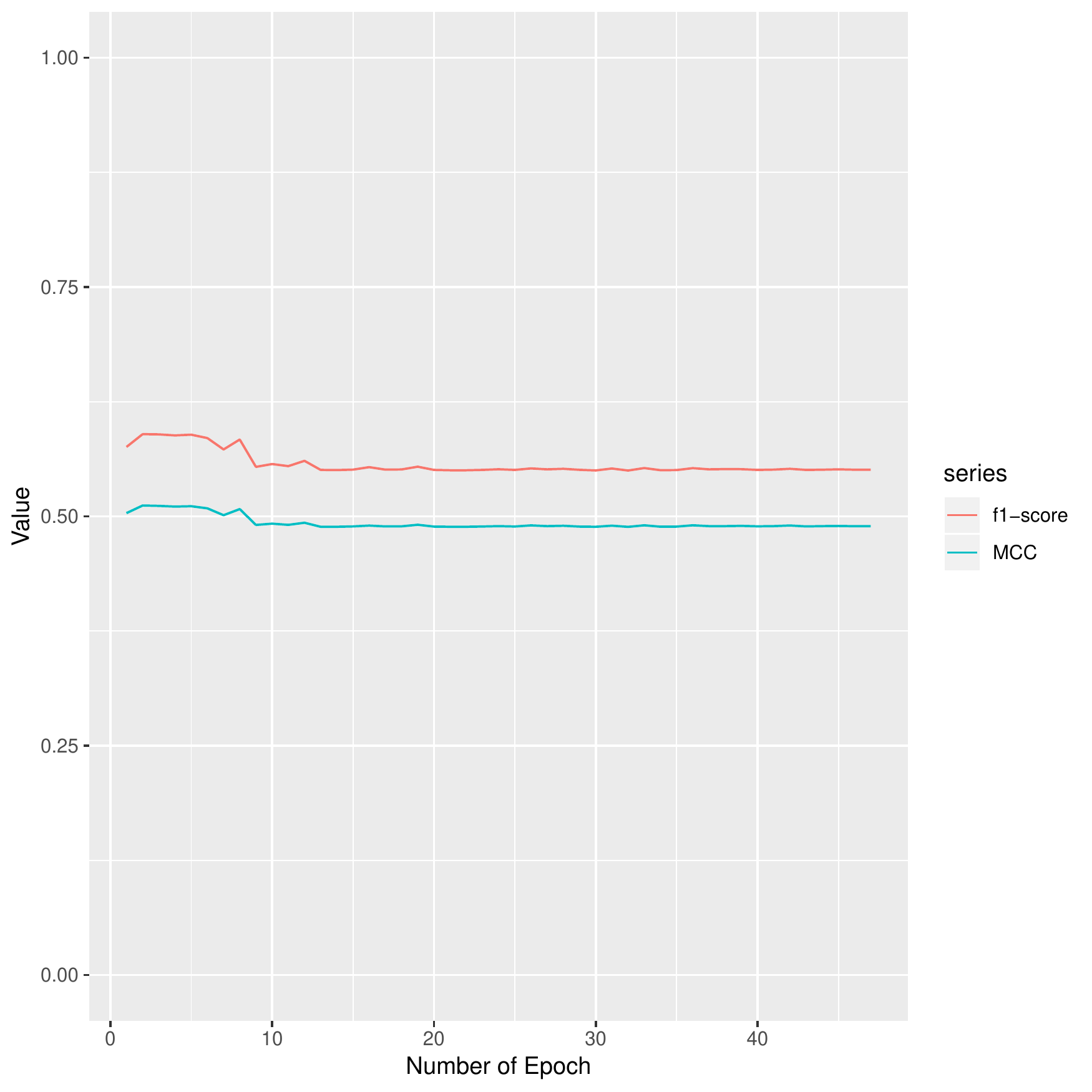}}
	\caption{Performance evaluation of our proposed method on different metric in Biological Process (BP) dataset.}
	\label{fig_4}
\end{figure}
\begin{figure}
	\centerline{\includegraphics[width = 75mm]{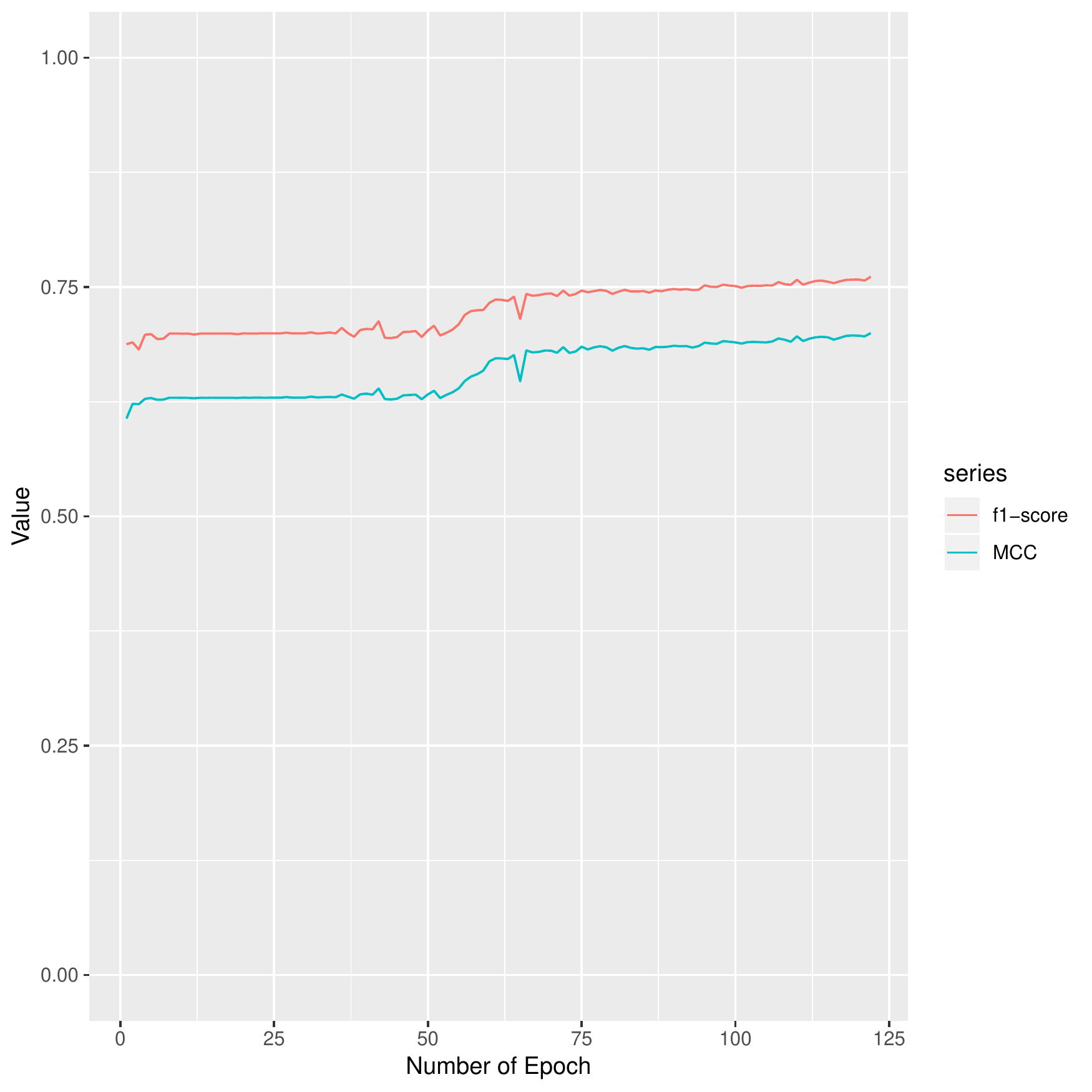}}
	\caption{Performance evaluation of our proposed method on different metric in Cellular Component dataset (CC).}
	\label{fig_5}
\end{figure}
Table~\ref{result}, shows the performance of our proposed DEEPGONET for predicting protein annotation. We have achieved impressive $F_{1}$-score $0.731, 0.763$ and $0.590$ in MF, CC, BP domains respectively. We also have achieved astonishing score in Mathew Correlation Coefficient (MCC), which are $0.706,0.700$ and $0.511$ in MF, CC, BP domains respectively. Fig.~\ref{fig_3},~\ref{fig_4},~\ref{fig_5} shows the performance of our models in  $F_{1}$-score and $MCC$ for MF, BP, CC dataset.

\subsection{Experimental Result on Various Organisms}
In order to test how much generalization our DEEPGONET has achieved, we tested our network to predict GO annotation in three domains for various organisms. The Table~\ref{model_species} shows the results.

\begin{table}
	\caption{DEEPGONET Network Performance on Various Organisms.}
	\begin{center}
		\begin{tabular}{cccc}			
			\hline \rule{0pt}{3ex}
			MF & \multicolumn{3}{c}{
				\begin{tabular}{ccc}
					\textbf{Organism} & \textbf{$F_{1}$-score} & \textbf{MCC}\\ \hline
					\textit{Arabidopsis thaliana} & 0.662 & 0.63\\ \hline
				    Zebrafish	& 0.591	& 0.562 \\ \hline
				    Rat	& 0.685	& 0.652 \\ \hline
				    \textit{Mus musculus} &	0.691 &	0.66  \\ \hline
				    \textit{Escherichia coli}  &	0.722 &	0.692 \\ \hline	
				    \textit{Pseudomonas aeruginosa} &	0.62 &	0.589 \\ \hline	
				    \textit{Bacillus subtilis} &	0.638 &	0.609 \\ \hline		
				    \textit{Mycobacterium tuberculosis} &	0.721 &	0.694 \\ [0.4mm]
				\end{tabular}		
			}\\ 
			
			\hline \rule{0pt}{3ex}
			CC& \multicolumn{3}{c}{
				\begin{tabular}{ccc}
					\textbf{Organism} & \textbf{$F_{1}$-score} & \textbf{MCC}\\ \hline
					\textit{Arabidopsis thaliana} & 0.784 & 0.693 \\ \hline
					Zebrafish	& 0.729	& 0.670 \\ \hline
					Rat	& 0.685	& 0.658 \\ \hline
					\textit{Mus musculus} &	0.711 &	0.637  \\ \hline
					\textit{Escherichia coli} &	0.670 &	0.623 \\ \hline	
					\textit{Pseudomonas aeruginosa} &	0.660 &	0.619 \\ \hline	
					\textit{Bacillus subtilis} &	0.706 &	0.672 \\ \hline		
					\textit{Mycobacterium tuberculosis} &	0.636 &	0.577 \\ [0.4mm]
				\end{tabular}		
			}\\ 
			
			\hline \rule{0pt}{3ex}
			BP & \multicolumn{3}{c}{
				\begin{tabular}{ccc}
					\textbf{Organism} & \textbf{$F_{1}$-score} & \textbf{MCC}\\ \hline
					\textit{Arabidopsis thaliana} & 0.526 & 0.460 \\ \hline
					Zebrafish	& 0.511	& 0.428 \\ \hline
					Rat	& 0.548	& 0.464 \\ \hline
					\textit{Mus musculus} &	0.569 &	0.491  \\ \hline
					\textit{Escherichia coli}  &	0.527 &	0.460 \\ \hline	
					\textit{Pseudomonas aeruginosa} &	0.540 &	0.519 \\ \hline	
					\textit{Bacillus subtilis} &	0.554 &	0.538 \\ \hline		
					\textit{Mycobacterium tuberculosis} &	0.560 &	0.521 \\ [0.4mm]
				\end{tabular}		
			}\\ \hline
			
		\end{tabular}
		\label{model_species}
	\end{center}
\end{table}

\section{Conclusions} \label{con}
In this paper, we present a novel architecture, DEEPGONET, for predicting top-level of GO ontology hierarchy in three domains. We trained our network using protein annotation from \textit{Homo sapiens} species only and tested it on various other organisms with impressive performance. With high performance measures in classification, DEEPGONET also allows the accurate automated annotation of proteins with respect to top-level of GO ontology hierarchy in all three domains by only modeling primary sequence of protein. Besides this, our DEEPGONET, to our best knowledge, first time implemented cascaded convolutional and recurrent neural network in GO ontology annotation prediction. Another contribution of this paper is the development of deep hierarchical classification model which optimizes predictive performance on top-level of hierarchies, and learns features in a hierarchical manner by accounting for dependencies among various classes. It is optimized concurrently through learning features in end-to-end fashion while using only primary sequence as input. This work has potential for being used in other application with similar structured output and can lead to paradigm shift in GO ontology experimental research because, instead of investing huge money and resource to explore all the branches of hierarchy, one can efficiently explore function of an protein by investigation the predicted branch(s) of the GO ontology hierarchy by our models.

So, in future, we intend to further extend this cascaded convolutional and recurrent hierarchical model in several directions. First, we plan to extend our model for predicting second top-level hierarchy of GO Ontology. We also plan to use multi-model input which includes different type of interactions such as protein-protein interaction, genetic interaction, co-expression network, and regulatory network.

\vspace{12pt}

\end{document}